\documentclass[11pt]{article}
\usepackage{graphics}
\usepackage[margin=1in]{geometry}

\usepackage{amssymb}
\usepackage{amsmath}
\usepackage{amsthm}
\usepackage{hyperref}
\usepackage{multicol}
\usepackage{booktabs}
\usepackage[ruled,linesnumbered,noend]{algorithm2e}

\newtheorem*{remark*}{Remark}

\title{
{\bf Large Neighborhood Search based on\\
	 Neural Construction Heuristics}}

\author{{\bf Jonas K. Falkner, Daniela Thyssens and Lars Schmidt-Thieme}\\
Department of Computer Science\\
University of Hildesheim, Germany\\
\\
Email of Corresponding Author: falkner@ismll.de}

\date{\today}

\begin{document}
\pagestyle{empty}

\maketitle
\thispagestyle{empty}

\noindent
{\bf Abstract}: 
We propose a Large Neighborhood Search (LNS) approach utilizing a learned construction heuristic based on neural networks as repair operator to solve the vehicle routing problem with time windows (VRPTW). Our method uses graph neural networks to encode the problem and auto-regressively decodes a solution and is trained with reinforcement learning on the construction task without requiring any labels for supervision.
The neural repair operator is combined with a local search routine, heuristic destruction operators and a selection procedure applied to a small population to arrive at a sophisticated solution approach. The key idea is to use the learned model to re-construct the partially destructed solution and to introduce randomness via the destruction heuristics (or the stochastic policy itself) to effectively explore a large neighborhood.
\vspace{0.5in}

\noindent
Team name:	 \textbf{JAMPR} \\
Solver name: \textbf{JAMPR} \\
VRP tracks:	 VRPTW \\
Code source: \url{https://github.com/jokofa/JAMPR_plus} \\

\pagebreak

%
%
%
%

\section{Neural Construction Heuristic}\label{sec:1}


The main feature and key novelty of our approach is the utilization of a neural construction heuristic (NCH) as repair operator in the LNS. Neural construction heuristics involve a neural network model, which is trained using the \textit{Learning-to-Optimize} paradigm. The paradigm consists of finding solutions to a problem, given, we have seen and solved many such problems (from the same distribution of problems) already, without having access to the optimal solutions themselves. Given a sampler for problem instances $p \sim \mathcal{P}$ and a cost function $\operatorname{cost}()$, learn a solver $\hat s$ that minimizes the cost of solutions for future problems from the same distribution:
\begin{equation}\label{eq:l2o}
\min \ \mathbb{E}_{p \sim \mathcal{P}}\left[\operatorname{cost}(\hat s(p)) \right].
\end{equation}

\subsection{Original model}
Our model and the training routine are heavily based on JAMPR\cite{jampr}. The original JAMPR model is an extension of the attention model (AM)\cite{kool} of Kool et al.\ which uses an encoder-decoder architecture involving self-attention\cite{vaswani}. Both models treat the Learning-to-Optimize problem as a sequential decision making problem, that is modeled as a Markov Decision Process and solved with reinforcement learning. The solution is constructed piece by piece by creating routes one node at a time, i.e.\ interpreting the current solution, route and unvisited nodes as state and the index of all unvisited nodes, which can feasibly be added to the current route, as actions.

First, the encoder receives the node features $x_i$ of each node $i$ (coordinates, demand, TW, etc.) and encodes them into a latent embedding vector $\tilde{x}_i \in \mathbb{R}^{d_{\text{emb}}}$ with dimension $d_{\text{emb}}$. Then the decoder model computes an attention query of each $\tilde{x}_i$ w.r.t.\ a specific context $C^{(t)}$ at decoding step $t$ to produce scores for all nodes which can feasibly be added to the current route. Here the context includes a latent embedding of the problem graph and additional problem information like the index of the depot node, the last node added to the current route and the remaining capacity. The resulting scores are then either used in a greedy selection procedure, i.e.\ always choosing the node with the highest score, or a softmax is employed to transform it into a distribution which is used for sampling. In general the encoder-decoder model represents the policy $\pi \left( i^{(t+1)} \mid C^{(t)}, x\ ; \theta\right)$ with learnable parameters $\theta$. 

JAMPR extended the standard AM, which was mainly developed for the TSP and CVRP, with additional encoders for routes and vehicles to enrich the context in order to tackle the VRPTW. For that purpose JAMPR creates a latent embedding for each constructed route $r \in R$ by feeding the node embeddings $\tilde{x}_i, i \in r$ and vehicle features $\psi_r$ (remaining capacity, current node, current time, etc.) of the corresponding route $r$ into additional fully connected neural networks, aggregating the outputs and concatenating them with the context. Since the thereby created extended context constitutes a more comprehensive state representation, it enables the parallel construction of several routes, which was found to be essential to successfully solve highly constrained problems like the VRPTW. The number of such concurrently planned routes is fixed to a constant $\kappa$.
This leads to a new extended action space of currently active routes $R^{(t)}_{\kappa}$ and available nodes equal to $A = \{(r, i) \in \text{feasible}(R^{(t)}_{\kappa} \times N)\}$ where $N$ is the number of customers and feasible() is a function selecting only nodes which can be added to the routes without violating any constraints.

\subsection{Adaptions and extensions}
While AM and JAMPR showed promising results on small instances with 20 to 100 customer nodes, the model performance seems to rapidly deteriorate for larger instances and out-of-distribution samples. Furthermore, the self-attention procedure considers the relation of each node to \textit{every} other node resulting in a complexity of $N^2$. 
This heavily impacts the computational burden in our case where we want to execute the construction procedure many times within an LNS framework. Moreover, for the VRPTW in particular, the distances in euclidean coordinates can differ from the incurred travel times, which can be difficult to learn. For that reason we introduce several major adaptions to the original JAMPR model:

\begin{itemize}
	\item We replace the self-attention layers with graph neural networks (GNN)\cite{gnn} which can work with explicit edge weights, e.g.\ the travel times in case of the VRPTW.
	
	\item We apply the GNN only to a spatial neighborhood $\mathcal{H}_i$ of each customer node $i$ which consists of a fixed number $k$ of nearest neighbors, thereby reducing the complexity from $N^2$ to $Nk$.	
	Since these neighborhoods do not change we can pre-compute static node embeddings $\tilde{x}_i^{\text{stat}}$ in the first forward pass and keep them fixed for the complete decoding procedure:
	\begin{equation}\label{eq:gnn_stat}
		\tilde{x}_i^{\text{stat}} = \operatorname{GNN}_{\text{stat}}\left(\mathcal{H}_i^{\text{stat}}\right).
	\end{equation}
	Stacking several such GNN layers increases the spatial resolution of the model w.r.t.\ the neighborhoods. 
	
	\item We add an additional GNN which creates a dynamic node embbedding based on the neighborhood $\mathcal{H}_{r^{(t)}, i}$ of route $r$ and node $i$ at step $t$, i.e.\ the incoming and outgoing edge of node $i=2$ in route $r^{(t)}=(0, 2, 7, 5)$ would be the set of edges $\{(0,2), (2,7)\}$. Accordingly:
	\begin{equation}\label{eq:gnn_dyn}
		\tilde{x}_i^{\text{dyn}} = \operatorname{GNN}_{\text{dyn}}\left(\mathcal{H}^{\text{dyn}}_{r^{(t)}, i}\right).
	\end{equation}
	The final node embedding is then created by simple summation of the embeedings: \\
	$\tilde{x}_i^{(t)} = \tilde{x}_i^{\text{stat}} + \tilde{x}_i^{\text{dyn}}$.
	
	\item We also introduce neighborhoods to the decoding procedure to reduce the size of the original action space of JAMPR. 
	We define the action neighborhood of route $r$ as $\mathcal{H}^{\text{a}}_{r^{(t)}} = \mathcal{H}_j \cup \{0\}$, where $j$ is the last node in route $r^{(t)}$ and 0 is the depot. Then the new action space is given by
	$A' = \{(r, i) \in \text{feasible}(R^{(t)}_{\kappa} \times \mathcal{H}^{\text{a}}_{r\in R^{(t)}_{\kappa}}) \}$.
	
	\item Finally, we follow POMO\cite{pomo} in replacing the original stochastic sampling procedure based on the softmax scores with a greedy decoding of multiple trajectories, which are created by sampling different configurations of start nodes for routes. To make the approach work with TW and multiple parallel routes we sort the nodes in the depot neighborhood $\mathcal{H}_0$ by the TW start time $t_i$ and randomly sample $\kappa$ start nodes from the 25\% earliest. Accordingly, we finally also change the original rollout baseline of \cite{kool} to the mean reward of the POMO trajectories.
	
\end{itemize}


\section{Large Neighborhood Search}
Our LNS follows the standard procedure described in \cite{lns} by employing dedicated repair and destruction operators, but further extends it with a local search routine and by maintaining a small population of candidate solutions. The complete method is given in algorithm \ref{alg:lns}. In contrast to our construction heuristic which considers one node at a time, Hottung and Tierney\cite{nlns} directly learn a neural repair operator to recombine tour fragments resulting from random splits for the CVRP. 

In the first iteration the pre-trained NCH is used to fully construct a set $M$ of $m_{\text{init}}$ feasible initial solutions. Out of this set we select a fixed number of samples. Namely we select the $m_{\text{best}}$ solutions with the smallest cost and $m_{\text{dissimilar}}$ solutions, which are the most dissimilar to the best solutions, leading to a population $Y$ of size $m = m_{\text{best}} + m_{\text{dissimilar}}$.
For the calculation of route similarity we treat routes as sequences and use sequence matching to calculate the similarity score as $\tau = 2\mu/L,\ \tau \in [0, 1]$ where $L$ is the total number of elements in both sequences and $\mu$ is the number of matches.
In the next step we execute a local search on all samples in $Y$ leading to $m$ possibly improved solutions which replace the samples in $Y$. After the local search we identify the new best solution $y^*$. If $y^*$ is not in $Y$, we replace the worst solution in $Y$ with $y^*$.
Then we apply the destruction operators and destroy each solution in $Y$ several times to arrive at the new diversified population $Y'$, which now is again of size $m_{\text{init}}$. Finally, the partial solutions $Y'$ are fed back into the NCH model, which reconstructs the solutions forming a new set $M$, and the process is repeated until some stop criterion is met.

\SetKwComment{Comment}{// }{ }
\SetKwFunction{SELECT}{SELECT}
\SetKwFunction{NCH}{NCH}
\SetKwFunction{LS}{LocalSearch}
\SetKwFunction{cost}{cost}
\SetKwFunction{getbest}{get\_best}
\SetKwFunction{replaceworst}{replace\_worst}
\SetKwFunction{DESTRUCT}{DESTRUCT}
\SetKwInOut{Input}{input}
\DontPrintSemicolon
\begin{algorithm}[ht]
	\caption{LNS with NCH}\label{alg:lns}
	
	\Input{Problem instance $x$, pre-trained \NCH, 
	$m_{\text{init}}$, $m_{\text{best}}$, $m_{\text{dissimilar}}$}
	
	$M \gets \NCH{$x, m_{\text{init}}$}$  \Comment*[r]{Create set of $m_{\text{init}}$ initial solutions}
	$y^* \gets \getbest{$M$}$ \\
	
	\Repeat{stop criterion is met}{	
		$Y \gets \SELECT{$M, m_{\text{best}}, m_{\text{dissimilar}}$}$   \Comment*[r]{get promising samples to form population}
		$Y \gets \LS{$x, Y$}$ \\
		$y \gets \getbest{$Y$}$ 		\Comment*[r]{identify candidate solution}
		\If{$\cost{$y$} < \cost{$y^*$}$}{
			$y^* \gets y$
		}
		\Else{ 
			$Y \gets \replaceworst{$Y, y^*$}$       \Comment*[r]{$y^* \notin Y$}
		}
		$Y' \gets \DESTRUCT{$Y, m_{\text{init}}$}$ 	\Comment*[r]{results in new set of $m_{\text{init}}$ partial solutions}
		$M \gets \NCH{$x, Y'$}$ 	\Comment*[r]{repair (re-construct)}
	}
	\Return{$y^*$}
\end{algorithm}

\subsection{Repair operator}
As repair operator we utilize the neural construction heuristic explained in section \ref{sec:1}. 
We train the NCH on the pure construction task for instances of size $N=50$. To accommodate some distribution shift we train four different models for the data modalities of type 1 (R1, C1, RC1) and type 2 (R2, C2, RC2) instances as well as for a low fraction (0.25-0.5) and high fraction (0.75-1.0) of TW. 
In the LNS the NCH receives a set of different partial solutions resulting from the destruction procedure and re-constructs new complete solutions. 
An important limitation is that in the current setup it can only reconstruct, i.e. repair, a maximum of $\kappa$ partial routes (it can always construct an arbitrary number of \textit{new} routes).

\subsection{Destruction operators}
The destruction operators in the LNS destroy parts of the solution to explore its neighborhood and escape local optima. Here we use a heuristic approach targeting different parts of the solution. We split the destruction into two different operations:

\paragraph{Creation of partial routes}
Since the maximum amount of partial routes is fixed to $\kappa$, we randomly select $\kappa$ routes $r$ and randomly apply one of four partial destruction operators: 
\begin{enumerate}
	\item \textit{random}: remove a fixed number of random nodes from the route,
	\item \textit{random\_cut}: remove a random number of nodes from the end of the route,
	\item \textit{const\_cut}: remove a fixed number of nodes from the end of the route,
	\item \textit{waiting\_time\_cut}: remove the node with the highest waiting time and all nodes thereafter.
\end{enumerate}

\paragraph{Removal of complete routes}
Furthermore, we can remove complete routes from the solution. Removal here means that all edges of the route are removed and all nodes are marked as unvisited. We propose a set of another four operators:
\begin{enumerate}
	\item \textit{random}: remove a random route from the solution,
	\item \textit{smallest}: remove the smallest route, i.e.\ the route with the smallest number of nodes,
	\item \textit{similar}: given a pair of solutions, take one as reference and remove the most similar route from the other solution using similarity score $\tau$,
	\item \textit{waiting\_time}: remove the route with the highest cumulative waiting time.
\end{enumerate}

\subsection{Local Search}
The local search is implemented using the Guided Local Search (GLS) method of the Google OR-Tools library \cite{ortools}.
Besides the standard search parameters for the GLS, we enable further local search operators, such as cross exchange and the relocation of chains of neighbor nodes. 
The time limit of the local search is set to 2-4 seconds for smaller problem sizes up to $N=200$ and 16-128 seconds for the remaining larger problem sizes.
Furthermore, we add penalty terms in cases where the maximal number of vehicles is succeeded. 

\section{Evaluation}
We report the results of our solver in the challenge VRPTW-track of the DIMACS VRP Challenge\footnote{\url{http://dimacs.rutgers.edu/programs/challenge/vrp/}} in table \ref{tab:results}. Our solver is able to achieve a cost of maximally 1.1 times the best known solution (BKS) for the majority of instances with up to 600 customer nodes. 

\begin{table}[ht]
\caption{Results of our solver for different instance types on the Solomon and Homberger \& Gehring benchmarks. We report the average and median PI as evaluated with the DIMACS VRP challenge controller provided by the organizers.}
\label{tab:results}
\centering
\small
\begin{tabular}{rlcccccc|cccccc} 
\toprule
\textbf{N} & & \multicolumn{2}{c}{\textbf{R1}} & \multicolumn{2}{c}{\textbf{C1}} & \multicolumn{2}{c}{\textbf{RC1}} & \multicolumn{2}{c}{\textbf{R2}} & \multicolumn{2}{c}{\textbf{C2}} & \multicolumn{2}{c}{\textbf{RC2}}\\
& \texttt{PI} &\texttt{avg} & \texttt{med} & \texttt{avg} & \texttt{med} & \texttt{avg} & \texttt{med} & \texttt{avg} & \texttt{med} & \texttt{avg} & \texttt{med} & \texttt{avg} & \texttt{med} \\
\hline
\hline  
100 &   & 7.069     & 7.565     & 6.447     & 7.014     & 8.748     & 9.498   
        & 7.708     & 9.161     & 9.141     & 9.702     & 8.264     & 9.699 \\ 

200 &   & 8.875     & 8.874     & 7.172     & 7.171     & 8.374     & 8.528   
        & 8.287     & 8.672     & 9.407     & 9.998     & 9.474     & 9.990 \\ 

400 &   & 8.745     & 9.293     & 8.359     & 9.427     & 9.409     & 9.634   
        & 8.328     & 8.539     & 9.488     & 10.0      & 8.955     & 9.533 \\ 

600 &   & 9.821     & 9.992     & 9.120     & 9.915     & 9.990     & 10.0   
        & 9.958     & 9.968     & 9.996     & 10.0      & 9.942     & 9.959 \\ 

        

\bottomrule
\end{tabular}
\end{table}

\section{Conclusion}

In this paper we propose a specialized LNS approach which utilizes a pre-trained neural construction heuristic. The experiments and challenge results show that the method is viable and, while being in a rather experimental stage, still shows promising performance and can compete (to some extent) with highly optimized algorithms, which in some cases are already in development for several years. 
Apart from that, there are still multiple parts of the algorithm which have high potential for improvement in future iterations. The local search can be adapted to be more focused on the specific task and the general interaction with the NCH and the LNS. The destruction heuristics are very simple and currently applied rather naively in a random fashion. Instead, more advanced rules could be used and their selection could be improved with an Adaptive LNS\cite{alns} or they could be replaced completely with a learned operator similar to the repair operator. Finally, the NCH itself can still be improved by explicit learning (or fine-tuning) for the reconstruction task in combination with the destruction operators instead of the current isolated pre-training.

\begin{remark*}
	We want to note that the evaluation setup of a \ {\normalfont single CPU - single thread} \ environment puts major limitations on our model. We argue that, while other algorithms can be parallelized as well, our NCH model is naturally parallelized on a GPU without any additional work and the whole surrounding computational framework (PyTorch) is tailored and optimized towards massive GPU parallelization compared to single CPU execution. For example, already a mid-segment consumer GPU (Nvidia GForce 1080ti) leads to an average $15$-$20\times$ speedup of the NCH.
\end{remark*}



\begin{thebibliography}{99}
	

\bibitem{jampr}
Falkner, J. K., \& Schmidt-Thieme, L. (2020). Learning to solve vehicle routing problems with time windows through joint attention. arXiv preprint arXiv:2006.09100.

\bibitem{nlns}
Hottung, A., \& Tierney, K. (2019). Neural large neighborhood search for the capacitated vehicle routing problem. arXiv preprint arXiv:1911.09539.

\bibitem{kool}
Kool, W., Van Hoof, H., \& Welling, M. (2018). Attention, learn to solve routing problems!. arXiv preprint arXiv:1803.08475.

\bibitem{pomo}
Kwon, Y. D., Choo, J., Kim, B., Yoon, I., Gwon, Y., \& Min, S. (2020). POMO: Policy Optimization with Multiple Optima for Reinforcement Learning. arXiv preprint arXiv:2010.16011.

\bibitem{gnn}
Morris, C., Ritzert, M., Fey, M., Hamilton, W. L., Lenssen, J. E., Rattan, G., \& Grohe, M. (2019, July). Weisfeiler and leman go neural: Higher-order graph neural networks. In Proceedings of the AAAI Conference on Artificial Intelligence (Vol. 33, No. 01, pp. 4602-4609).

\bibitem{ortools}
Perron, L. and Furnon, V. (2019, July). Or-tools (https://developers.google.com/optimization/).

\bibitem{lns}
Pisinger, D., \& Ropke, S. (2010). Large neighborhood search. In Handbook of metaheuristics (pp. 399-419). Springer, Boston, MA.

\bibitem{alns}
Ropke, S., \& Pisinger, D. (2006). An adaptive large neighborhood search heuristic for the pickup and delivery problem with time windows. Transportation science, 40(4), 455-472.

\bibitem{vaswani}
Vaswani, A., Shazeer, N., Parmar, N., Uszkoreit, J., Jones, L., Gomez, A. N., ... \& Polosukhin, I. (2017). Attention is all you need. In Advances in neural information processing systems (pp. 5998-6008).


\end{thebibliography}
\end{document}